\pdfoutput=1

\documentclass[11pt]{article}

\usepackage[final]{acl}

\usepackage{times}
\usepackage{latexsym}

\usepackage[T1]{fontenc}

\usepackage[utf8]{inputenc}
\usepackage{tcolorbox}
\usepackage{enumitem}
\usepackage{microtype}

\usepackage{amsmath}
\usepackage{inconsolata}

\usepackage{graphicx}

\usepackage[capitalize,noabbrev]{cleveref}
\usepackage{booktabs}
\usepackage{multirow}

\def \VersionWithComments {}
\ifdefined \VersionWithComments
\usepackage{marginnote}
\newcommand{\marginX}{\marginnote{\huge{\quad\quad\textbf{!}\quad\quad}}}

\newcommand{\todo}[1]{\mbox{}{\color{blue}{\marginX{}\textbf{TODO}\ifx#1\\\else:\ \fi #1}}}
\fi

\title{\textbf{ReMedi}: Reasoner for Medical Clinical Prediction}

\author{
 \textbf{Yushi Cao\textsuperscript{1}},
 \textbf{Yiming Chen\textsuperscript{1}}\thanks{Corresponding author.},
 \textbf{Hongchao Jiang\textsuperscript{1}},
 \textbf{Hung-yi Lee\textsuperscript{2}},
 \textbf{Robby T. Tan\textsuperscript{1,3}} 
\\
 \textsuperscript{1}ASUS Intelligent Cloud Services (AICS), Singapore\\
 \textsuperscript{2}NTU Artificial Intelligence Center of Research Excellence (NTU AI-CoRE), Taiwan\\
 \textsuperscript{3}National University of Singapore, Singapore
\\
\tt yushi\_cao@asus.com \quad Hongchao\_Jiang@asus.com \quad MattYM\_Chen@asus.com \\
\tt  hungyilee@ntu.edu.tw \quad robby.tan@nus.edu.sg
}

\begin{document}
\maketitle
\begin{abstract}
Predicting future clinical outcomes from electronic health records (EHR) remains challenging due to the complexity and heterogeneity of patient data.
LLMs have shown strong potential for such predictive tasks, yet existing approaches mainly focus on enhancing medical knowledge through distillation or RAG while relying on the model's internal ability to interpret contextual information.
In this work, we present \textbf{ReMedi} (\underline{Re}asoner for \underline{Me}dical Clinical Pre\underline{di}ction), a framework for improving clinical outcome prediction from EHR. 
\textbf{ReMedi} generates rationale–answer pairs using a challenging sample re-generation mechanism for complex clinical questions, which leverages ground-truth answers as hints to enhance reasoning for further supervised fine-tuning and preference tuning. \textbf{ReMedi} integrates ground-truth outcome guidance into the preference data construction loop, regenerating rationale-answer variants. By tuning on these rationale-answer pairs, the model improves its predictive performance on clinical prediction tasks.
Experiments on multiple EHR prediction tasks demonstrate substantial gains of up to $19.9\%$ over state-of-the-art baselines in terms of F1 score, underscoring \textbf{ReMedi}'s effectiveness in real-world clinical prediction.

%

%

\end{abstract}

\section{Introduction}
\label{sec:intro}
Predicting future outcomes such as readmission, length of stay, and mortality is crucial for improving patient care and hospital resource management.
Electronic Health Records (EHRs)~\citep{mimic3,johnson2023mimic4,wornow2023ehrshot} provide rich longitudinal data but are difficult to model due to their structured and heterogeneous nature.
While Large Language Models (LLMs) show promise in processing medical data~\citep{clmber1,UniRep,medbert}, they still struggle to interpret EHR inputs effectively.\footnote{See APPX.~\ref{appx:related-works} for a detailed discussion of related works.}

Despite recent progress in clinical prediction using LLMs~\citep{xu2024ram,graphcare}, most approaches underutilize the models' reasoning abilities, which have proven effective in complex problem-solving domains~\citep{deepseekr1,wang2025baichuan,wu2025medreason}. 
For instance, KARE~\citep{kare} incorporates reasoning through knowledge distillation and structured medical graphs, but its reliance on proprietary teacher models and predefined ontologies hinders scalability and real-world deployment.

In this work, we propose \textbf{ReMedi} (\underline{Re}asoner for \underline{Me}dical Clinical Pre\underline{di}ction), a simple and effective framework that improves clinical outcome prediction through self-generated supervised fine-tuning (SFT) and Direct Preference Optimization (DPO)~\citep{rafailov2023direct}. 
\textbf{ReMedi} leverages a challenging sample re-generation process to utilize complex and difficult clinical prediction questions for training data construction.
We further devise \textbf{iReMedi}, an iterative variant of ReMedi, which follows an iterative refinement process that progressively enhances the predictive accuracy across multiple training rounds.

We evaluate the proposed \textbf{ReMedi} on three clinical prediction tasks, where it significantly outperforms existing methods.
By incorporating iterative training, \textbf{iReMedi} further improves overall performance.
Notably, we observe that directly applying existing general-domain self-improvement approaches (e.g., STaR~\citep{zelikman2022star}) to clinical prediction tasks yields only marginal gains due to training inefficiency.
Finally, we conduct ablation studies to examine the contribution of each sub-component.


\begin{figure*}[t!]
\centering
  \includegraphics[width=\linewidth]{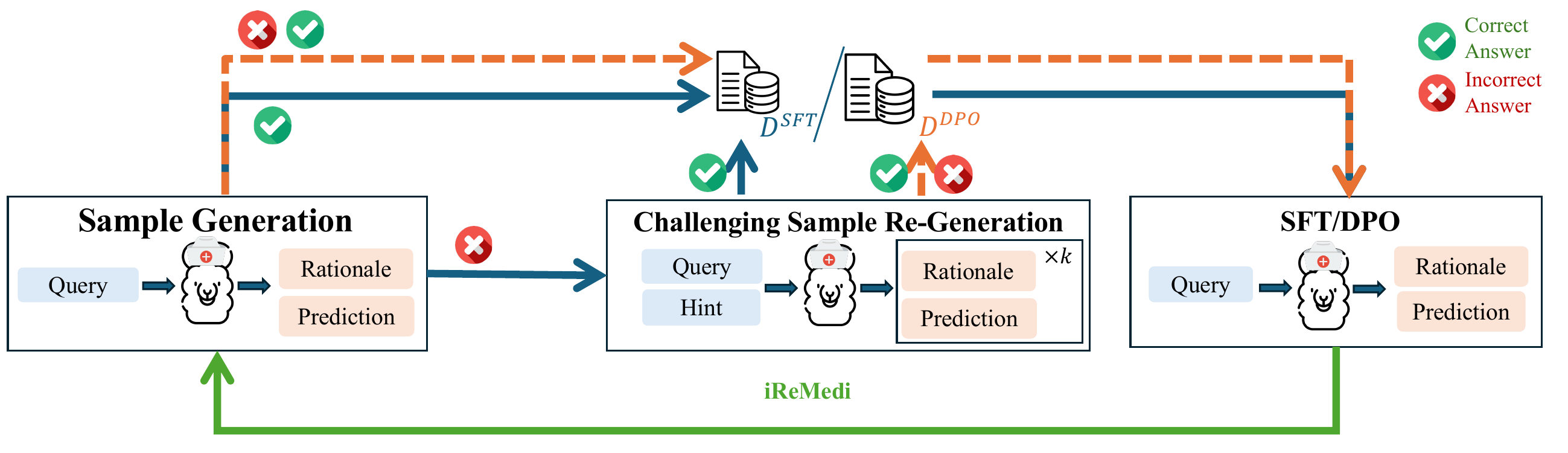}
  \caption{Overview of \textbf{ReMedi}, which operates iteratively across three stages: (1) Sample Generation, (2) Challenging Sample Re-Generation, and (3) Model Training. The dotted orange line represents the data processing pipeline for DPO, while the solid blue line denotes the pipeline for SFT.}
  \label{fig:methodology:overview}
\end{figure*}

Our main contributions are as follows:
%
(1) We propose \textbf{ReMedi}, an efficient and effective framework for clinical prediction tasks. 
\textbf{ReMedi} introduces outcome-guided preference construction, integrating ground-truth outcomes into the generation of rationale–answer pairs and forming an improvement process optimized via SFT and DPO.
(2) \textbf{ReMedi} incorporates challenging sample re-generation to focus learning on difficult clinical prediction tasks.
(3) Experiments on three standard EHR benchmarks show that ReMedi outperforms existing methods by up to $19.9\%$ in terms of F1 score, and extensive ablation analyses validate its advantages in terms of predictive performance.

\section{ReMedi}
\paragraph{Overview} The overview of \textbf{ReMedi} is illustrated in~\Cref{fig:methodology:overview}.
\textbf{ReMedi} enhances clinical prediction through a three-stage framework:
(1) Sample Generation,
(2) Challenging Sample Re-Generation, and
(3) Model Training.
We further introduce an iterative variant, \textbf{iReMedi}, which repeatedly executes these three stages to progressively refine the model’s reasoning and prediction ability.

\subsection{Sample Generation }
The \textbf{ReMedi} cycle begins with the sample generation stage. 
Given an initial dataset $D = \{(q_i, a_i)\}_{i=1}^N$, where $q_i$ denotes a query, $a_i$ its ground-truth answer, and $N$ the total number of samples, the generator model $M$ is prompted to produce rationale–answer pairs: $\{(\hat{r}_i, \hat{a}_i) \sim M(q_i)\}_{i=1}^N$.
Since ground-truth answers are available, we can evaluate and filter the generated pairs. 
Intuitively, high-quality rationales tend to produce correct answers.
We retain those pairs where the generated answer matches the ground truth to construct a supervised fine-tuning dataset: 
$D^{\text{SFT}} = \{(q_i, \hat{r}_i, \hat{a}_i) \mid \hat{a}_i = a_i,\, q_i \in D\}$.
To improve data efficiency, we also collect rationale–answer pairs with incorrect predictions and pair them with correct ones to form the DPO dataset:
$D^{\text{DPO}} = \{(q_i, \hat{r}_i, \hat{a}_i, \hat{r}_i', \hat{a}_i') \mid \hat{a}_i = a_i,\, \hat{a}_i' \neq a_i,\, q_i \in D\}$.

To ensure a rapid adaptation to clinical prediction tasks, we leverage a warm-start (WS) phase for the initial model generator $M$. Specifically, during this phase, we directly apply both label rationalization and repeated sampling to all queries, and then gather correct rationale-answer pairs to ensure the quality of the generated data. 

\subsection{Challenging Sample Re-Generation }
Queries that the model answers incorrectly typically represent challenging cases, and leveraging such samples has been shown to significantly improve model performance~\citep{muennighoff2025s1}. 
To effectively utilize these difficult cases, we employ two strategies: \textit{label rationalization} and \textit{repeated sampling}.
Label rationalization differs from standard answer generation by providing the model with additional grounding. 
During this process, the model receives both the original query and the ground-truth label as hints, helping it generate more coherent rationales, particularly in the early stages when domain knowledge is still limited.

To further maximize the value of these challenging queries, we generate $k$ responses per query and select one correct rationale–answer pair from the multiple samples. 
However, this approach risks the model explicitly referencing the hint during reasoning. To mitigate this, we filter out any rationale that explicitly mentions the given hint. 
Samples that consistently fail to produce the correct answer, even when hints are provided, are discarded. The remaining rationale–answer pairs are incorporated into $D^{\text{SFT}}$ for supervised fine-tuning, while $D^{\text{DPO}}$ is augmented with the newly generated, more challenging samples.

\subsection{Model Training}
After generating the synthetic data, we fine-tune the model in two stages: SFT and DPO. 
First, the base model $M$ is fine-tuned on the supervised dataset $D_{\text{SFT}}$ to obtain the SFT model. 
Next, the SFT model is used to collect preference data, forming the dataset $D_{\text{DPO}}$, which is then employed to further optimize the model through DPO training built upon the SFT model.
The model optimization follows standard SFT and DPO objectives~\citep{rafailov2023direct}, 
where SFT minimizes the cross-entropy loss over correct rationale–answer pairs, 
and DPO optimizes the log-preference ratio between preferred and dispreferred samples.


\subsection{Iterative ReMedi (iReMedi)}
Building on the success of iterative training for improving LLM reasoning, we propose \textbf{iReMedi}, an iterative extension of \textbf{ReMedi}.
In each iteration, the model undergoes the complete ReMedi optimization pipeline to progressively enhance both reasoning quality and predictive accuracy.
The updated model $M^*$ then serves as the generator for the next round of sample generation.
To mitigate overfitting and maintain generalization, the training process is reinitialized from the original base model $M$ in each ReMedi round.

\begin{table*}[!htbp]
\centering
\resizebox{\textwidth}{!}{
\begin{tabular}{clcccccccccc} 
\toprule
 \multirow{2}{*}{\textbf{Type}} & \multirow{2}{*}{\textbf{Method}} & \multicolumn{4}{c}{\textbf{Mortality Prediction}} & \multicolumn{4}{c}{\textbf{Readmission Prediction}} & \multicolumn{2}{c}{\textbf{Length of Stay}} \\ 
\cmidrule(lr){3-6} \cmidrule(lr){7-10} \cmidrule(lr){11-12}
 & & \textbf{Acc.} & \textbf{F1} & \textbf{TPR.} & \textbf{TNR.} & \textbf{Acc.} & \textbf{F1} & \textbf{TPR.} & \textbf{TNR.} & \textbf{Acc.} & \textbf{F1} \\ 
\midrule

\multirow{13}{*}{Prompting} & Direct Prediction & & & & & & & & \\
 & ~~ Gemini-2.5-Flash & 49.3  & 33.7 & 99.6 & 0.80 & 62.8  & 62.6 & 94.1  & 44.0  & 31.7& 23.6 \\
 & ~~ HuotuoGPT-o1-72B & 61.5 & 60.6  & 62.5  & 60.0 & 61.0 & 55.9 & 77.8 & 34.8 & 39.8 & 25.9 \\ 
 \cmidrule{2-12}
 & Zero-shot (CoT) & & & & & & & & \\
 & ~~ Gemini-2.5-Flash & 68.0 & 67.9 & 83.5 & 58.9 & 50.3 & 36.1 & \colorbox{lightgray}{99.2} & 3.12 &41.5&36.6 \\
 & ~~ HuotuoGPT-o1-7B & 48.2 & 44.6 & 97.6 & 18.3 & 49.9 & 36.8 & 97.1 & 4.35 &26.3&16.4 \\ 
 & ~~ MedReason & 37.4 & 28.0 & \colorbox{lightgray}{100.0} & 1.00 & 50.1 & 35.6 & 98.3 & 2.68 & 27.3&25.1 \\ 
 & ~~ MedGemma & 63.8 & 63.8 & 89.6 & 50.7 & 50.1 & 33.6 & 99.2 & 0.26 & 23.6 & 18.7 \\
\cmidrule{2-12}
 & Few-shot (CoT) & & & & & & & & \\
 & ~~ Gemini-2.5-Flash & 71.3 & 71.0 & 82.7 & 64.5 & 49.8 & 35.1 & \colorbox{lightgray}{99.2} & 2.21 &43.9 &38.7 \\
 & ~~ HuotuoGPT-o1-7B & 75.2 & 73.9 & 78.6 & 73.5 & 52.2 & 41.8 & 96.8 & 9.73 & 31.4 &24.6 \\ 
 & ~~ MedReason & 45.6 & 41.8 & 96.6 & 15.9 & 49.9 & 36.6 & 95.1 & 4.18 &29.7 &24.6 \\ 
  & ~~ MedGemma & 51.4 & 44.6 & 55.6 & 50.7 & 51.5 & 39.2 & 98.2 & 6.71 &29.4 & 20.2 \\
\midrule
 \multirow{4}{*}{Fine-tuning} & ~~SFT & 88.9 & 88.3 & 82.8 & 92.9 & 69.2 & 66.4 & 91.4 & 43.5 &39.9&36.6\\
 & ~~KARE & 95.9 & 95.5 & \colorbox{lightgray}{95.1} & 97.5 & 81.2 & 81.3 & 83.7 & 78.3 &40.4 &35.9 \\
 & ~~ReMedi (Ours) & 97.7\small{(+1.8)} & \colorbox{lightgray}{97.6}\small{(+2.1)} & 94.3\small{(-0.8)} & \colorbox{lightgray}{100.0}\small{(+2.5)} & 90.5\small{(+9.3)} & 90.4\small{(+9.1)} & 80.6\small{(-3.1)} & \colorbox{lightgray}{100.0}\small{(+21.7)} &55.6\small{(+15.2)} &55.5\small{(+19.6)} \\
 & ~~iReMedi (Ours) & \colorbox{lightgray}{97.8}\small{(+1.9)} & \colorbox{lightgray}{97.6}\small{(+2.1)} & 94.1\small{(-1.0)} & \colorbox{lightgray}{100.0}\small{(+2.5)} & \colorbox{lightgray}{91.5}\small{(+10.3)} & \colorbox{lightgray}{91.4}\small{(+10.3)} & 83.8\small{(+0.1)} & \colorbox{lightgray}{100.0}\small{(+21.7)} & \colorbox{lightgray}{56.1}\small{(+15.7)} & \colorbox{lightgray}{55.8}\small{(+19.9)} \\
\bottomrule
\end{tabular}
}

\caption{Comparisons between ReMedi and baselines. Following~\citep{kare,graphcare}, we report the Accuracy, Macro F1, True Positive Rate (TPR), and True Negative Rate (TNR). The best performance is \colorbox{lightgray}{highlighted}. The values in brackets are the absolute performance gain between our method and KARE.}
\label{tab:mainresult}
\end{table*}

\section{Experiments}

\subsection{Experimental Setup}

\paragraph{Implementation Details}

We leverage HuatuoGPT-o1 as the base model to utilize its enormous medical knowledge learned through its pre-training and post-training stages, reducing the need for complex medical knowledge retrieval~\citep{kare,graphcare}.
We fine-tune our model using the TRL~\citep{vonwerra2022trl}, Tranformers~\citep{wolf-etal-2020-transformers}), DeepSpeed~\citep{rasley2020deepspeed}, and Flash-Attention2~\citep{dao2023flashattention2} frameworks.
Following~\citep{kare}, for all the fine-tuning models we use set the initial learning rate to 5e-6 using AdamW~\citep{adamw} optimizer with a batch size of 16.
We follow~\citep{kare} for data processing. We use the Clinical Classifications Software to map the medical codes of conditions and procedures to natural language and use the Anatomical Therapeutic Chemical system for medication. The train, test, and validation sets are split in a 0.8/0.1/0.1 ratio.


\paragraph{Dataset and Tasks}
We process the publicly available MIMIC-IV EHR dataset~\cite{johnson2023mimic4} into three standard clinical prediction tasks following~\cite{kare,graphcare,johnson2023mimic4}:
(1) \textit{Mortality Prediction}: predicting whether a patient will die during the next hospital visit;
(2) \textit{Readmission Prediction}: predicting whether the patient will be readmitted within 15 days after discharge;
(3) \textit{Length of Stay}: predicting the duration of hospitalization for the current visit.
We follow~\citep{kare} for data processing. More specifically, we retain 10,000 samples for the readmission task (with 5,000 samples labeled as readmission, and 5,000 as no-readmission). For mortality prediction, 2,701 patients with a mortality outcome and 7,299 patients with a survival outcome are retained. For Length of Stay, we set four classes: less than one day, one to seven days, one to two weeks, and more than two weeks. We obtain a total number of 10,000 samples (with 2,500 samples for each class).

\paragraph{Annotation Instruction}
Given the generated rationale-answer pairs, we check if the content in the rationale matches the prediction results. For example, if the rationale thinks the patient will be re-admitted, then the prediction should be $1$ (readmission), otherwise, it is treated as a misalignment.

\paragraph{Baselines} 
We compare against two categories of methods:
(1) \textit{Prompting-based LLMs}: Zero-shot and Few-shot prompting (with reasoning trace) with Gemini-2.5-Flash~\cite{team2023gemini}, HuatuoGPT-o1-7B~\cite{chen2024huatuogpt}, MedReason~\cite{wu2025medreason}, and MedGemma-27B~\cite{sellergren2025medgemma}; Direct prediction (without reasoning trace) with Gemini-2.5-Flash and HuatuoGPT-o1-72B.
(2) \textit{Fine-tuned LLMs}: KARE~\cite{kare} and a HuatuoGPT-o1-7B variant trained directly on ground-truth data.

\subsection{Main Results}
\Cref{tab:mainresult} summarizes the main results of \textbf{ReMedi} compared with all baselines. We observe that \textbf{ReMedi} and \textbf{iReMedi} consistently achieve the best performance across all tasks. Notably, all prompting-based methods exhibit near-random guessing behavior, reflecting their inability to produce high-quality data for clinical prediction tasks unfamiliar to general-purpose LLMs. This highlights the difficulty of clinical prediction tasks.

\textbf{ReMedi} outperforms both fine-tuning baselines, KARE and the SFT variant of HuatuoGPT-o1-7B (that is, fine-tuning on ground-truth data once only). For mortality and readmission prediction, \textbf{ReMedi} achieves absolute accuracy improvements of $1.8\%$ and $9.1\%$, respectively, compared to KARE. For the length-of-stay task, it yields larger gains of $15.2\%$ in accuracy and $19.6\%$ in F1 score. Additionally, \textbf{iReMedi} further outperforms ReMedi and other baselines, which demonstrates the effectiveness of the three-stage learning framework.

\subsection{Discussion}


\paragraph{Effect of Individual Components}
We conduct an ablation study to evaluate the contribution of each component to \textbf{ReMedi}'s performance. Additionally, we also consider the iterative self-taught learning method (STaR~\citep{zelikman2022star}).
Results of each setting are summarized in~\Cref{tab:ablation}. 
Incorporating DPO training leads to consistent performance gains for both \textbf{ReMedi} and \textbf{iReMedi}.
Compared with STaR, the proposed \textbf{iReMedi} achieves markedly better results, primarily attributed to the challenging sample re-generation component, which effectively utilize the complex and challenging samples.


\begin{table}[ht]
\centering
\resizebox{\columnwidth}{!}{
\begin{tabular}{lcccc} 
\toprule 
                  \textbf{Method}               & \textbf{Acc.} & \textbf{F1}   & \textbf{TPR.} & \textbf{TNR.}  \\ 
\midrule
ReMedi        & 90.5 & 90.4 & 80.6 & \colorbox{lightgray}{100.0}   \\
ReMedi w/o DPO & 84.4 & 84.4 & 85.3 & 83.6   \\
iReMedi       & \colorbox{lightgray}{91.5} & \colorbox{lightgray}{91.4} & 83.8 & \colorbox{lightgray}{100.0}   \\
iReMedi w/o DPO & 86.8 & 86.8 & 83.7 & 89.9 \\
STaR  & 59.1 & 53.2 & \colorbox{lightgray}{96.1} & 23.4   \\ 
\bottomrule
\end{tabular}
}
\caption{Effect of different sub-components. The best performance is \colorbox{lightgray}{highlighted}.}
\label{tab:ablation}
\end{table}


\paragraph{Alignment Between Thinking and Prediction}
We manually evaluated responses from both KARE and the proposed \textbf{ReMedi} to examine the consistency between reasoning processes and final predictions. Specifically, 40 instances were randomly sampled from each of the readmission and no-readmission classes. 
Additionally, we utilize Gemini-2.5-Flash to evaluate the extent of misalignment across all test samples, and the results are consistent with the manual evaluation.
The results, summarized in Table~\ref{tab:alignment}, show that for the no-readmission class, both KARE and \textbf{ReMedi} exhibit strong consistency between reasoning and prediction. However, in the readmission class, both methods display varying degrees of misalignment.
Notably, KARE suffers from more severe misalignment, likely due to its multi-task learning design~\citep{kare}, where reasoning generation and label prediction are trained jointly.
In contrast, \textbf{ReMedi}, with its end-to-end learning scheme, produces rationale–answer pairs that are better aligned and more coherent, leading to reduced misalignment.
Nonetheless, occasional misalignment may still arise during data generation and filtering.


\begin{table}
\centering
\resizebox{\columnwidth}{!}{
\begin{tabular}{ccccccc} 
\toprule
\multirow{2}{*}{Method} & \multicolumn{2}{c}{\textbf{No Readmission}} & \multicolumn{2}{c}{~\textbf{Readmission}} & \multicolumn{2}{c}{\textbf{Avg.}}  \\ 
\cmidrule{2-7}
                        & Human & Gemini           & Human & Gemini         & Human & Gemini  \\ 
\midrule
KARE                    & 25.0  & 14.0                       & 95.0  & 90.0                     & 60.0  & 52.0              \\
ReMedi                  & 85.0  & 80.6                       & 100.0 & 100.0                    & 92.5  & 90.0              \\
\bottomrule
\end{tabular}
}
\caption{Alignment between thinking and prediction.}
\label{tab:alignment}
\end{table}


\section{Case Study}
To investigate why certain cases are particularly challenging, we conducted a case study involving three patients who were admitted to the hospital due to appendicitis. The differences are that: Patient A has no other chronic diseases (no readmission); Patient B has asthma (no readmission); Patient C has cancer (readmission). The difficulty lies in how to differentiate the nuanced clinical judgment about different conditions. The HuatuoGPT-o1 can predict Patient A and C correctly, but incorrectly predicts Patient B as readmission. It means that existing LLMs, such as HuatuoGPT-o1, do not differentiate the subtle distinctions in risk assessment and overestimate the risks (e.g., typical appendectomy recovery and stable chronic conditions). This also aligns with the results in Table~\ref{tab:mainresult}, where the TPR (True Positive Rate) is high and the TNR (True Negative Rate) is low. By contrast, after fine-tuning, ReMedi demonstrates the capacity to perform more structured clinical reasoning. It evaluates risk by integrating both the severity of individual symptoms and their typical clinical trajectories. This suggests that ReMedi has learned to make finer-grained risk assessments by aligning observed symptoms and treatments with expected outcomes, thereby distinguishing between high-risk and low-risk cases with greater nuance.

Huatuo Response to Patient B: "....With the chronic condition, the potential for future hospital visits is definitely there. Plus, the appendectomy itself might have introduced new health concerns or complications... So, considering all this, the patient's appendectomy and recovery stage, along with the asthma, I'm leaning towards the prediction that this patient might be readmitted within 15 days."

ReMedi Response to Patient B: "... the most prominent factor here is the recent appendectomy. It requires close monitoring, and any complications could lead to readmission. The patient's current condition and medication regimen suggest they are managing well. Asthma could be a potential risk factor, but there's no clear indication of an imminent exacerbation as no specific procedures or medication are taken for asthma.... Therefore, the likelihood of readmission is not high."

\section{Conclusion}
We present \textbf{ReMedi}, a framework for improving clinical outcome prediction from EHR. 
\textbf{ReMedi} integrates supervised fine-tuning and DPO, leveraging the challenging sample re-generation mechanism, where complex clinical prediction questions are effectively utilized. This design enhances both predictive accuracy and task adaptability across diverse clinical settings. 
Experiments on multiple EHR benchmarks show that \textbf{ReMedi} achieves substantial gains over state-of-the-art methods. 
For future work, we intend to explore extending \textbf{ReMedi} to more challenging, open-question clinical tasks as well as multimodal medical prediction tasks.


\section*{Limitations}
Despite the promising results achieved by \textbf{ReMedi}, several limitations warrant consideration. 
First, \textbf{ReMedi} exhibits minor misalignment between the rationale and the prediction. How to further mitigate such misalignment towards a coherent rationale and prediction needs to be investigated. 
Second, the assessment focuses on well-established clinical prediction tasks, such as mortality and hospital readmission, which have definitive outcome labels. The capability of the model (and the self-consistency strategy) to handle more complex, open-ended clinical reasoning tasks remains an open question. 
Third, this work primarily leverages relatively small-scale language models (e.g., HuatuoGPT-o1-7B); the performance and scalability of the approach when applied to significantly larger models (e.g., those exceeding 70 billion parameters) have yet to be systematically evaluated. 
Lastly, due to the limited availability of medical experts, the human evaluation is restricted to assessing the alignment between the model’s reasoning and its final prediction. We acknowledge that rigorously verifying the clinical correctness and validity of the reasoning process with domain experts is an important direction for future work.

\section*{Ethical Considerations}
An important ethical consideration arises from the use of sensitive electronic health record (EHR) data obtained from the MIMIC-IV dataset. To ensure responsible data handling, we have completed the required Data Use Agreement\footnote{https://physionet.org/about/licenses/physionet-credentialed-health-data-license-150/} and strictly adhere to its terms and conditions. The Gemini-2.5-Flash is accessed via Vertex, which is approved by the MIMIC4 dataset provider\footnote{https://physionet.org/news/post/gpt-responsible-use}.
Furthermore, all model training was conducted locally using secured, in-house computational resources, thereby minimizing potential risks associated with data privacy and external data transmission.


\bibliography{anthology,custom}

\appendix

\section{Related Works}
\label{appx:related-works}
\paragraph{Traditional Clinical Prediction Models}
Electronic health record (EHR) data contains a vast amount of detailed patient information (normally stored as medical codes), making it a valuable source in the medical and clinical domains~\cite{johnson2023mimic4,mimic3,wornow2023ehrshot}. With the development of deep learning, researchers started to learn and capture the complex latent patterns of the EHR data~\cite{clmber1, UniRep,pang2021cehrbert}. Methods like CLMBER~\cite{clmber1,clmber2} utilize sophisticated network structures to learn the latent representations of the EHR data for various downstream clinical prediction tasks~\cite{choi2016retain,choi2017gram,gao2020stagenet}. Another line of methods focuses on constructing graphs to link all the information centered on the patient, aiming to improve prediction accuracy~\cite{graphcare,graph1,graph2}. However, such traditional models are often inflexible and require specific construction on the training and input data, which is inadequate for the dynamically changing healthcare domain.

\paragraph{LLM for Clinical Predictions}
LLMs~\cite{gpt4,llama3,deepseekr1} have demonstrated strong capabilities in various domains~\citep{jeong-etal-2024-medical,chen-etal-2024-beyond-single,MEDSAGE,jiang2025codejudgebench,li2024llms}.
Motivated by this, researchers are actively harnessing LLMs for precise clinical prediction tasks. Most approaches utilize retrieval-augmented generation (RAG) to obtain useful external knowledge from public medical databases and construct prompts together with patient information~\cite{xu2024ram,zhu2024emerge,ye2021medretriever,niu2024ehr}. To create more specific and high-quality prompts, Jiang et al.~\cite{kare} construct knowledge graphs from external medical knowledge tailored by patient information to create more fine-grained and precise content. Such methods mainly focus on retrieving related information and constructing augmented input for prompting or reasoning distillation from larger models. Therefore, the model's intrinsic reasoning capabilities are often overlooked. In contrast, our approach focuses on fully harnessing recent medical reasoning models' internal reasoning abilities with little to no access to larger models.

\paragraph{Reasoning Enhancement}
With the evolution of large language models (LLMs), a variety of approaches have been proposed to enhance their reasoning capabilities.
One prominent direction is Self-improvement, which is closely related to self-taught learning. This approach aims to enhance the reasoning capabilities of large language models (LLMs) through the use of synthetic data, thereby reducing dependence on ground-truth labels~\citep{huang2023large,chen2024self,singhbeyond,burns2024weak}. This paradigm primarily relies on the ability of powerful, off-the-shelf LLMs to evaluate or select candidate answers—often through techniques such as majority voting—to improve output quality~\citep{huang2023large}. However, the reliability of these judgments is not guaranteed and may inadvertently reinforce incorrect or biased behavior if the underlying model is flawed~\citep{raina2024llm,cheng2024self2}. In our work, we provide ground truth labels as hints so that the reasoning behavior and final prediction are aligned and guided by the ground truth, mitigating the potential intrinsic behavior/bias of the LLMs.

Self-correction has also shown promise in improving the quality of LLMs' outputs. Many previous works~\citep{shinn2023reflexion,huanglarge,kim2023language,miao2023selfcheck} employ multiple rounds of feedback and correction to correct their own outputs. However, as noted in~\citep{huanglarge}, current LLMs continue to face challenges in identifying and correcting errors without external feedback. Moreover, these models may convert correct responses into incorrect ones during the self-correction process. Additionally, self-correction typically incurs longer inference times due to the need for multiple rounds of internal evaluation. In contrast, our approach prompts the model to generate both the reasoning trace and the final answer in a single pass, eliminating the need for iterative feedback and correction mechanisms.

Bootstrapping is a widely adopted strategy for enhancing the reasoning capabilities of large language models (LLMs). It typically involves generating new training samples using the model itself, which are then used to improve performance through supervised fine-tuning (SFT) or reinforcement learning (RL)~\citep{yuan2023scalingrelationshiplearningmathematical,dong2023raft,zelikman2022star,guo2025deepseek,ulmer2024bootstrapping,zhang2023bootstrap}. When ground-truth labels are available, rejection sampling is often employed to filter out high-quality data for post-training~\citep{yuan2023scalingrelationshiplearningmathematical,dong2023raft,zelikman2022star,guo2025deepseek}. In scenarios where ground-truth labels are unavailable, many approaches depend on existing models to either generate data or guide its refinement~\citep{ulmer2024bootstrapping,zhang2023bootstrap}. In our work, we improve the efficiency of the bootstrapping process by including a warm-up stage. This is especially useful when the initial models are not well-adapted to the task (e.g., clinical prediction using EHR data).

\paragraph{Comparison between ReMedi and STaR}
The distinctions between ReMedi and STaR can be summarized into three key aspects:
\begin{itemize}
    \item \textbf{Pipeline design:} ReMedi’s main method is not iterative. It performs a single SFT stage followed by DPO. STaR, in contrast, depends on iterative self-training loops as its primary mechanism. Our iterative variant (iReMedi) was added, yielding marginal improvements (90.5->91.5), suggesting that the main performance gains arise from the design of ReMedi itself.
    \item \textbf{Data construction and supervision:} In STaR, the model first attempts to generate a rationale-answer pair without hints; for the examples it still gets wrong, a hint is then provided, and the model regenerates a rationale-answer pair. Only those pairs that ultimately yield the correct answer are retained, while unsuccessful attempts are discarded. ReMedi differs in two key ways: (i) for each challenging query, we generate  candidate rationale-prediction pairs and select one usable pair to maximize the value of each challenging query; and (ii) during our DPO stage we leverage both correct and incorrect rationale-prediction pairs rather than training only on the successful ones, thus providing a richer/informative supervision signal.
    \item \textbf{Clinical-domain adaptation:} We acknowledge the effectiveness of self-improvement frameworks demonstrated in prior work. However, directly applying STaR-like strategies to clinical prediction tasks is non-trivial. As shown in Table 2, a direct application of the STaR framework yields only 59.1\% accuracy—far below the performance achieved by ReMedi. This highlights that the success of clinical reasoning tasks requires more than simply transferring existing self-training methods. This motivates our key design choices, which are tailored to novel tasks like clinical prediction tasks and are essential for strong performance.
\end{itemize}

\section{Prompts}

\begin{tcolorbox}[title=Prompt with ground truth hint, fonttitle=\scriptsize]
\scriptsize
\texttt{Given the following task description, patient EHR context, please provide a step-by-step reasoning process that leads to the prediction outcome based on the patient's context.}

\texttt{After the reasoning process, provide the prediction strictly follow this format:}

\texttt{\# Prediction \# Your prediction}

\texttt{======================================================}

\texttt{\# Task \#}

\texttt{Readmission Prediction Task:}

\texttt{Objective: Predict if the patient will be readmitted to the hospital within 15 days of discharge.}

\texttt{Labels: 1 = readmission within 15 days, 0 = no readmission within 15 days}

\texttt{Note: Analyze the information comprehensively to determine the likelihood of readmission. The goal is to accurately distinguish between patients who are likely to be readmitted and those who are not.}

\texttt{======================================================}

\texttt{\# Patient EHR Context \#}

\texttt{\textbf{[[Patient EHR]]}}

\texttt{======================================================}

\texttt{\# Ground Truth \#}

\texttt{\# Prediction \# \textbf{[[Ground Truth Hint]]}}

\texttt{======================================================}

\texttt{Please provide a step-by-step reasoning process that leads to the correct prediction based on the patient's context and the ground truth.}

\texttt{The reasoning should be comprehensive, medically sound, and clearly explain how the patient's information leads to the predicted outcome. Your reasoning process must align with the ground truth provided. You cannot mention the ground truth in your reasoning process.}
\newline

\texttt{Then, provide your final prediction label in this format:}

\texttt{\# Prediction \# Your answer}
\newline

\texttt{**Important Notes:**}

\texttt{- You must follow the ground truth to generate the reasoning process!!}

\texttt{- Pretend that you do not know about the ground truth and do not mention the ground truth label in the reasoning process!!}

\end{tcolorbox}

\begin{tcolorbox}[title=Prompt with ground truth hint, fonttitle=\scriptsize]
\scriptsize
\texttt{Given the following task description, patient EHR context, please provide a step-by-step reasoning process that leads to the prediction outcome based on the patient's context.}

\texttt{After the reasoning process, provide the prediction strictly follow this format:}

\texttt{\# Prediction \# Your prediction}

\texttt{======================================================}

\texttt{\# Task \#}

\texttt{Readmission Prediction Task:}

\texttt{Objective: Predict if the patient will be readmitted to the hospital within 15 days of discharge.}

\texttt{Labels: 1 = readmission within 15 days, 0 = no readmission within 15 days}

\texttt{Note: Analyze the information comprehensively to determine the likelihood of readmission. The goal is to accurately distinguish between patients who are likely to be readmitted and those who are not.}

\texttt{======================================================}

\texttt{\# Patient EHR Context \#}

\texttt{\textbf{[[Patient EHR]]}}

\texttt{======================================================}

\texttt{Please provide a step-by-step reasoning process that leads to the correct prediction based on the patient's context.}

\texttt{The reasoning should be comprehensive, medically sound, and clearly explain how the patient's information leads to the predicted outcome.}
\newline

\texttt{Then, provide your final prediction label in this format:}

\texttt{\# Prediction \# Your answer}

\end{tcolorbox}

\end{document}